\documentclass{article}

\usepackage[preprint]{neurips_2020}
\usepackage[utf8]{inputenc} % allow utf-8 input
\usepackage[T1]{fontenc}    % use 8-bit T1 fonts
\usepackage{hyperref}       % hyperlinks
\usepackage{url}            % simple URL typesetting
\usepackage{booktabs}       % professional-quality tables
\usepackage{amsfonts}       % blackboard math symbols
\usepackage{nicefrac}       % compact symbols for 1/2, etc.
\usepackage{microtype}      % microtypography
\usepackage{enumerate}
\usepackage{amsthm}

\newtheorem{theorem}{Theorem}[section]
\newtheorem{lemma}[theorem]{Lemma}

\newtheorem{definition}[theorem]{Definition}

\title{An improvement to a result about graph isomorphism networks using the prime factorization theorem}

\author{%
  Rahul Sarkar\\
  Institute for Computational and Mathematical Engineering\\
  Stanford University\\
  \texttt{rsarkar@stanford.edu} \\
}

\begin{document}

\maketitle

\begin{abstract}
The unique prime factorization theorem is used to show the existence of a function on a countable set $\mathcal{X}$ so that the sum aggregator function is injective on all multisets of $\mathcal{X}$ of finite size.
\end{abstract}

\section{Introduction}

In the seminal work \citep[Section~4]{xu2018how}, the authors introduce a class of Graph Neural Networks (GNNs) called \textit{Graph Isomorphism Networks} and prove some universality results about them. A key lemma that they prove which underlies their universality results is \citep[Lemma~5]{xu2018how}, which we restate below for convenience:

\begin{lemma}
\label{lem:injec-existence}
Let $\mathcal{X}$ be a countable set. Then there exists a function $f: \mathcal{X} \rightarrow \mathbb{R}^n$, so that the sum aggregator function $h(X) = \sum_{x \in X} f(x)$ is unique for each multiset $X \subset \mathcal{X}$ of bounded size. Moreover, any multiset function $g$ can be decomposed as $g(X) = \phi \left( \sum_{x \in X} f(x) \right)$ for some function $\phi$.
\end{lemma}

To the best of knowledge of the author, subsequent work on GNNs have not removed the boundedness condition on multisets in Lemma~\ref{lem:injec-existence}. In this short note, we use the unique prime factorization theorem to show that Lemma~\ref{lem:injec-existence} holds for all multisets of finite size.

\section{Main result}
\label{sec:main-result}

We will assume that $\mathcal{X}$ is a countable set. 

First we define the notion of multisets, which generalizes the concept of sets, allowing multiple occurrences of the same element.
\begin{definition}
\label{def:multisets}
A multiset $X$ of $\mathcal{X}$, is a $2$-tuple $X := (S,m)$ where $S$ is a subset of $\mathcal{X}$, called the underlying set, and $m: S \rightarrow \mathbb{N}_{\ge 1}$ is a multiplicity function that gives the multiplicity of the elements. We say that $X$ has finite size if $\sum_{x \in S} m(x)$ is finite, in which case we call this sum the size of $X$.
\end{definition}

We can now state and prove the main result of this paper. Denote by $\mathbb{N}_{p}$ the set of all primes, and let $\mathcal{Q}$ denote the set of all multisets of $\mathcal{X}$ of finite size. As $\mathcal{X}$ is countable it is clear that $\mathcal{Q}$ is also countable.

\begin{theorem}
\label{thm:main-result-1}
There exists a function $f: \mathcal{X} \rightarrow \mathbb{R}$, so that the function $h(X) = \sum_{x \in X} f(x)$ defined for each multiset $X \in \mathcal{Q}$, is an injective function on $\mathcal{Q}$, and any such function $f$ must be injective. Moreover, any multiset function $g:\mathcal{Q} \rightarrow \mathcal{Y}$, where $\mathcal{Y}$ is any arbitrary set, can be decomposed as $g(X) = (\phi \circ h) \left(X \right)$ for some function $\phi: \mathbb{R} \rightarrow \mathcal{Y}$.
\end{theorem}

\begin{proof}
We first prove the first statement. Assuming existence of $f$, injectivity of $f$ follows directly from the injectivity of $h$, as we obtain $f$ by restricting $h$ to multisets of size one. It remains to prove the existence of $f$. Since $\mathcal{X}$ is countable, let $\beta: \mathcal{X} \rightarrow \mathbb{N}_{p}$ be any bijective map. Define $f(x) := \log (\beta(x))$ for each $x \in \mathcal{X}$. We proceed via contradiction. Let $X_1 := (S_1,m_1)$ and $X_2 := (S_2,m_2)$ be two distinct finite sized multisets of $\mathcal{X}$, and suppose $h(X_1)=h(X_2)$. This implies $\sum_{x \in S_1} m_1(x) \log (\beta(x)) = \sum_{y \in S_2} m_2(y) \log (\beta(y))$, or equivalently
\begin{equation}
\label{eq:proof-main-result-1}
    \prod_{x \in S_1} \beta(x)^{m_1(x)} = \prod_{y \in S_2} \beta(y)^{m_2(y)}.
\end{equation}
By the prime factorization theorem, this implies $S_1 = S_2$ and $m_1 = m_2$, which contradicts the distinctness of $X_1$ and $X_2$.

The second part is trivial: define $\phi$ to be any function such that $\phi (h(X)) = g(X)$,  for any $X \in \mathcal{Q}$, and this is always well-defined for any arbitrary $g$ because $h$ is injective on $\mathcal{Q}$.
\end{proof}

We can use this theorem to prove a result analogous to \citep[Corollary~6]{xu2018how}, which we state as a theorem below and provide a proof. Notice that the $\epsilon$ parameter appearing in the statement of the theorem below is not allowed to be any irrational number, unlike  in \citep[Corollary~6]{xu2018how}.

\begin{theorem}
\label{thm:main-result-2}
There exists $f: \mathcal{X} \rightarrow \mathbb{R}$, and a set $Z \subset \mathbb{R}$ of Lebesgue measure zero, such that for any $\epsilon \in \mathbb{R} \setminus Z$, the function $\tilde{h}(c,X) := \epsilon f(c) + \sum_{x \in X} f(x)$ defined for each $2$-tuple $(c,X)$, where $c \in \mathcal{X}$ and $X \in \mathcal{Q}$, is an injective function on $\mathcal{X} \times \mathcal{Q}$. Any such function $f$ must be injective. Moreover, any function $\tilde{g}: \mathcal{X} \times \mathcal{Q} \rightarrow \mathcal{Y}$, where $\mathcal{Y}$ is any arbitrary set, can be decomposed as $\tilde{g}(c,X) = \phi \left( \tilde{h}(c,X) \right)$, for some function $\phi: \mathbb{R} \rightarrow \mathcal{Y}$.
\end{theorem}

\begin{proof}
Define the function $\gamma: \mathbb{Q}_{> 0} \times \mathbb{Q}_{> 0} \rightarrow \mathbb{R}$, as $\gamma(p,q) = \log_q (p)$. Let $Z$ be the image of $\gamma$. As $\mathbb{Q}_{> 0}$ is countable, being the set of positive rational numbers, so is the cartesian product $\mathbb{Q}_{> 0} \times \mathbb{Q}_{> 0}$, and hence $Z$ is also countable and has Lebesgue measure zero. Now let $f$ be defined as in the proof of Theorem~\ref{thm:main-result-1}, and choose any $\epsilon \in \mathbb{R} \setminus Z$. We again proceed via contradiction. Suppose $(c_1,X_1), \; (c_2,X_2) \in \mathcal{X} \times \mathcal{Q}$ be distinct elements, and assume $\tilde{h}(c_1,X_1) = \tilde{h}(c_2,X_2)$. Then there are two distinct cases which we need to consider separately: (a) $c_1 = c_2$ and $X_1 \ne X_2$, and (b) $c_1 \neq c_2$. If case (a) holds, then we have $\sum_{x \in X_1} f(x) = \sum_{x \in X_2} f(x)$, but then by Theorem~\ref{thm:main-result-1} we obtain that $X_1 = X_2$ which is a contradiction. If case (b) holds, then we get
\begin{equation}
\label{eq:proof-main-result-2}
    \epsilon \left( \log(\beta(c_1)) - \log(\beta(c_2)) \right) = \sum_{x \in X_2} \log(\beta(x)) - \sum_{x \in X_1} \log(\beta(x)).
\end{equation}
Now suppose $X_1 := (S_1,m_1)$ and $X_2 := (S_2,m_2)$. We immediately deduce from (\ref{eq:proof-main-result-2}) that 
\begin{equation}
\label{eq:proof-main-result-2-1}
    \left( \frac{\beta(c_1)}{\beta(c_2)} \right)^\epsilon = \frac{\prod_{x \in S_2} \beta(x)^{m_2(x)}}{\prod_{x \in S_1} \beta(x)^{m_1(x)}},
\end{equation}
and so $\epsilon \in Z$, which is again a contradiction. This proves the first statement.

It is clear that the second statement is true, because if $f(c_1) = f(c_2)$ for distinct elements $c_1,c_2 \in \mathcal{X}$, then for multisets $X_1$ and $X_2$ of size one defined on the underlying sets $\{c_1\}$ and $\{c_2\}$ respectively, we have $\tilde{h}(c,X_1)=\tilde{h}(c,X_2)$ for any $c \in \mathcal{X}$. This contradicts injectivity of $\tilde{h}$. The third statement is trivial: simply define $\phi$ to be any function such that $\phi(\tilde{h}(c,X)) = \tilde{g} (c,X)$, for any $2$-tuple $(c,X) \in \mathcal{X} \times \mathcal{Q}$, which is well-defined by injectivity of $\tilde{h}$.
\end{proof}

\bibliographystyle{plainnat}
\bibliography{bibliography}

\begin{thebibliography}{1}
\providecommand{\natexlab}[1]{#1}
\providecommand{\url}[1]{\texttt{#1}}
\expandafter\ifx\csname urlstyle\endcsname\relax
  \providecommand{\doi}[1]{doi: #1}\else
  \providecommand{\doi}{doi: \begingroup \urlstyle{rm}\Url}\fi

\bibitem[Xu et~al.(2019)Xu, Hu, Leskovec, and Jegelka]{xu2018how}
Keyulu Xu, Weihua Hu, Jure Leskovec, and Stefanie Jegelka.
\newblock How powerful are graph neural networks?
\newblock In \emph{International Conference on Learning Representations}, 2019.
\newblock URL \url{https://openreview.net/forum?id=ryGs6iA5Km}.

\end{thebibliography}

\end{document}